# Improved Partial Differential Equation and Fast Approximation Algorithm for Hazy/Underwater/Dust Storm Image Enhancement

**Uche A. Nnolim**
*Department of Electronic Engineering, Faculty of Engineering,*
*University of Nigeria, Nsukka, Enugu, Nigeria;*
uche.nnolim@unn.edu.ng

**ABSTRACT**

This paper presents an improved and modified partial differential equation (PDE)-based de-hazing algorithm. The proposed method combines logarithmic image processing models in a PDE formulation refined with linear filter-based operators in either spatial or frequency domain. Additionally, a fast, simplified de-hazing function approximation of the hazy image formation model is developed in combination with fuzzy homomorphic refinement. The proposed algorithm solves the problem of image darkening and over-enhancement of edges in addition to enhancement of dark image regions encountered in previous formulations. This is in addition to avoiding enhancement of sky regions in de-hazed images while avoiding halo effect. Furthermore, the proposed algorithm is utilized for underwater and dust storm image enhancement with the incorporation of a modified global contrast enhancement algorithm. Experimental comparisons indicate that the proposed approach surpasses a majority of the algorithms from the literature based on quantitative image quality metrics.

***Keywords*:** Logarithmic image processing; illumination-reflectance model; filter kernel-based enhancement; filter-based dynamic range compression; power law-based illumination correction; partial differential equations.

## 1  Introduction

Image scenes taken in outdoor environments can be affected by weather resulting in hazy appearance [1]. This leads to poor visibility as a result of scattering and absorption of light by suspended particles in the air [1]. Thus, there is the need for haze removal to improve image visibility and contrast in computer vision applications such as surveillance, intelligent transport systems, etc. Image de-hazing is a current and active area of research in image processing [2]. There are numerous approaches which are classified as single- or multi- image-based schemes [2]. Furthermore, interest has shifted to single image based methods due to feasibility and cost concerns inherent in multiple image-based methods.

The single image-based methods can be classified under enhancement, restoration, fusion- and deep-learning-based domains [2]. Previously, restoration-based approaches were the preferred route for image de-hazing with the dark channel prior (DCP)-based method [3] being the most popular and incorporated in numerous algorithms [1]. For instance, Li et al proposed an image de-hazing method involving content-adaptive dark channel employing an associative filter, with structure transfer ability for efficient dark channel computation [4]. This is in addition to post enhancement of the luminance of the de-hazed image and local contrast preservation [4]. Also, due to the excessive computation time of the Laplacian-based soft mapping procedure in the original DCP, He et al proposed the guided filter to reduce the runtime [5]. However, the guided filter does not preserve fine structures according to Li and Zheng [6]. Thus, they proposed a novel globally guided image filtering (G-GIF) composed of both global structure transfer filter and global edge-preserving smoothing filters for image fine structure preservation-based de-hazing [6].

Recently, fusion [7] [8] [9] [10] [11] and deep learning-based approaches [12] have become increasingly popular and widespread [2] due to faster and greater computing resources. However, a vast amount of images is required in the training stage in addition to considerable computing resources and runtime for the deep learning approaches. The fusion-based methods are relatively much more involved than the purely enhancement-based approaches but effective in most cases. Also, the restoration methods require tuning of several parameters for different images to obtain best results.

Image enhancement is usually employed as a pre-processing step in image processing and computer vision [13]. Additionally, the relatively recent application of partial differential equations (PDEs) for image enhancement and illumination correction has widened the possibilities and capabilities of proposed algorithms. PDE-based approaches are more versatile and flexible in addition to being able to yield intermediate and gradual results [14].

Additionally, PDE-based enhancement-based approaches have been applied to de-hazing and underwater image processing [15] [16] [17] [18]. Furthermore, based on previous works, it was shown that the phenomenon of illumination was similar to haze [17]. Thus, it followed that by reversing hazy images, illumination correction algorithms could be applied to process hazy images with some modifications. However, it was also discovered that only certain illumination correction algorithms were effective in de-hazing.

Though several problems inherent in the de-hazing processes have been addressed, other issues still persist. These include darkened images after de-hazing, over-enhancement of noise artifacts and halo effect, especially in restoration- and enhancement-based methods [17]. Though recent solutions appeared to solve the halo problem, the edge over-enhancement and image darkening became more pronounced [18]. Additional solutions were sought to mitigate these issues but did not lead to consistent and satisfactory results in all images. Thus, there is a need to address these issues without increasing complexity.

The proposed tonal mapping-based de-hazing algorithm solves the problem of darkened and over-enhanced de-hazed images by utilization of a log-less operator combined a with a haze approximation method defined in a PDE-based formulation. The scheme is optimized using the gradient-based metric after which the obtained image is refined using a filter-based enhancement operator. Further additions are incorporated for processing underwater and sandstorm images. Additional improvements involved edge-agnostic fuzzy logic-based enhancement for darkened de-hazed outlier images.

The outline of the paper is as follows; the second section presents a brief overview and background relating to illumination and reflectance estimation. The third section presents the proposed modified algorithm while the fourth section deals with the experiments and related results. The final section presents the conclusions.

## 2  Brief overview and background

The haze in images can be likened to illumination in dark or shadowy images [17] since the end result of uneven illumination and haze is the reduction in image visibility and contrast. Thus, by maximizing contrast, the visibility of the image scene is enhanced, reducing the haze or uneven illumination. This is manifested by edge enhancement or increase in gradients across the image scene. The same explanation is applicable to dusty and underwater images, which also exhibit haze and/or uneven illumination. Based on this realization, we can describe the example categories of dark, hazy and underwater images depicted in Fig. 1 as shown in Table 1. What all these image categories have in common is that edges or details are less pronounced and that they exhibit low visibility and contrast. The generalized enhancement models for processing these type of images is as shown in Fig. 1(b). The enhancement operator can be based on tonal correction, exponential, power law or statistical enhancement operators & previous works utilized these configurations for processing the afore-mentioned images types.

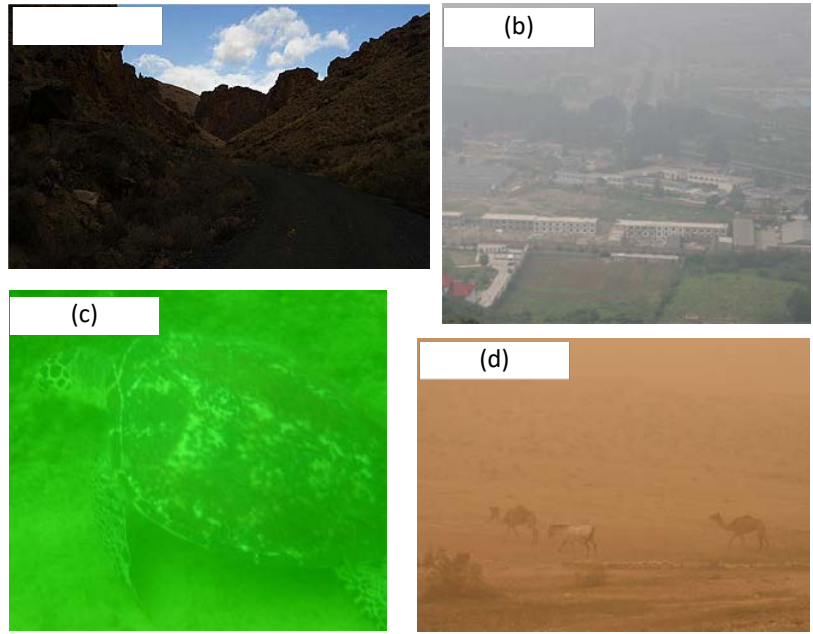

Figure 1: Examples of (a) dark, (b) hazy, (c) underwater and (d) sandstorm images

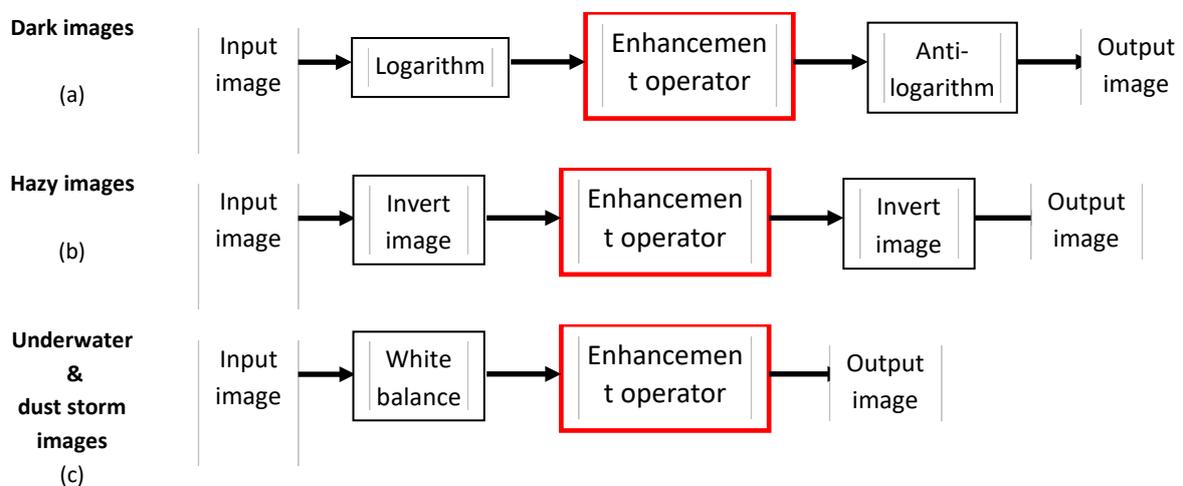

Figure 2: Generalized enhancement models for (a) dark, (b) hazy and (c) dust storm & underwater images

Table 1: features of dark, hazy, underwater and dust storm images

| Dark images | Hazy images | Underwater images | Dust storm images |
|---|---|---|---|
| Suffer from uneven illumination and shadow effects. | Suffer from haze and fog. | Suffer from haze, fog and green or blue haze/colour channel dominance/distortion | Suffer from haze due to dust. |
| Dark images with shadows | Images with poor visibility and low contrast with foggy appearance. | Images with colour channel dominance/tint. E.g. bluish or greenish appearance. | Images with brownish/reddish appearance. |
| Illumination varies slowly across image. | Haze varies slowly across image. | Can suffer from both poor illumination and haze. | Can also suffer from both poor illumination and haze. |

In summary, illumination and hazy conditions are similar in nature and underwater images may suffer from additional illumination or foggy conditions (or a combination of both) in addition to colour channel distortion. These issues are handled best with Retinex-based operators without modification. However, in the absence of Retinex, statistics-based contrast enhancement methods also yield reasonable results. The Retinex works for

- o Dark images (due to logarithm function)
- o Hazy images due to local and global (multiscale) surround functions for enhancement
- o Underwater images due to local and global enhancement properties.
- o Contributions of proposed approach include:
- o Repurposing of a log-less logarithmic image processing (LIP) algorithm for de-hazing in a PDE framework.
- o Modification of algorithm for underwater and sandstorm image enhancement.
- o Post refinement of de-hazing results using filter- and fuzzy logic-based enhancement.
- o Development of a fast de-hazing algorithm based on approximation of hazy formation model.

## 3 Proposed algorithm and modifications

The proposed approach seeks to solve the problems and shortcomings of the previously proposed PDE-based approaches [19] [18]. The former method performed remarkably well for illumination correction and de-hazing, but still required some improvements in terms of reduction of darkening of image regions in de-hazed images.

### 3.1 Revised PDE-based formulation

The PDE-based method performed relatively well in illumination correction but led to images with colour fading and darkened regions with over-sharpened edges when used for de-hazing applications [18]. This was mitigated by using frequency emphasis filters to increase brightness and preserving low frequency components, reducing noise due to over-sharpening [18]. Thus, based on research and experiments, gradient increase is a reliable indicator of image enhancement. Thus, the average gradient (AG) [20] was chosen as a suitable and reliable metric for the stopping criterion of the PDE and this modification was also adapted to the de-hazing process [18]. Though results were improved, the absence of local contrast enhancement and darkening of image regions still persisted in varying degrees.

#### 3.1.1 PDE-based formulation using Patrascu LIP for de-hazing

One of the drawbacks encountered in contrast enhancement of hazy images is the darkening of processed images. This was also true of the revised PDE formulation, leading to the exploration of other algorithms to address this problem. This was coupled with the challenge to maintain the relatively low-complexity of the previous algorithm and its advantages while improving its performance. Thus, in order to avoid complex and complicated algorithms, we selected the logarithmic image processing (LIP) formulation by Patrascu et al [21], which does not involve actual logarithmic calculation. In this section, we modify the proposed approach to utilize this algorithm, which does not affect edges. It should be noted that the hardware architecture of the modified LIP algorithm was also developed and verified in previous work [22]. This makes it a practical alternative for use in the formulation.

Positive attributes of the LIP included dynamic range compression without colour fading in addition to preservation of image highlights in bright images. The algorithm was subsequently adapted to image de-hazing but with initially mixed results. Thus it was reformulated into a PDE flow, incorporating aspects such as adaptive computation of the regularization parameter, $\alpha$ and AG-based optimization. Initially, the exponent, $\lambda$, was fixed for the LIP. However, we re-imagined this parameter as determined by the amount of dark/black area (BA) and light/white area (WA) in the images. By adaptively computing $\lambda$, we account for the differences in each hazy image. The LIP guides the evolution of the image in the PDE formulation and the results were better than the previous formulation [18]. However, some hazy images were still darkened or had over-enhanced sky regions, though the adaptive computation of $\lambda$ reduced the degree of the effect. Furthermore, the addition of the illumination-reflectance contrast enhancement scheme (IRCES) as a post-processing operation brightened the images. The PDE-based expression is given as;

$$\frac{\partial I(x,y,t)}{\partial t} = \alpha\big(f\{I(x,y,t)\} - I(x,y,t)\big) + \frac{\beta(I(x,y,t)-\mu)}{\sigma} + \Delta I(x,y,t) \qquad (1)$$

In (1), the enhancement operator is given as; $f\{I\} = \lambda <\times> I = \frac{[1+I]^\lambda - [1-I]^\lambda}{[1+I]^\lambda + [1-I]^\lambda}$ and $\lambda$ is the exponent. Substituting for $f\{I(x,y,t)\}$ leads to the expression in (2);

$$\frac{\partial I(x,y,t)}{\partial t} = \alpha\left(\frac{[1+I(x,y,t)]^\lambda - [1-I(x,y,t)]^\lambda}{[1+I(x,y,t)]^\lambda + [1-I(x,y,t)]^\lambda} - I(x,y,t)\right) + \frac{\beta(I(x,y,t)-\mu)}{\sigma} + \Delta I(x,y,t) \qquad (2)$$

Using the finite difference method (FDM) yields the discrete form as;

$$I^{t+1}(x,y) = I^t(x,y) + \left[\alpha\left(\frac{[1+I(x,y,t)]^\lambda - [1-I(x,y,t)]^\lambda}{[1+I(x,y,t)]^\lambda + [1-I(x,y,t)]^\lambda} - I(x,y,t)\right) + \frac{\beta(I(x,y,t)-\mu)}{\sigma} + \Delta I(x,y,t)\right]\Delta t \qquad (3)$$

After the processing, we now utilize the IRCES to brighten the image and enhance the edges without amplifying the noise as; $I_e(x,y) = IRCES(I(x,y))$. This complete process is the proposed algorithm (PA) and it solves the problem of dark images after de-hazing operations in most cases but may also lighten other ones. The approach is still amenable to hardware implementation and just requires the inversion of the image before and after processing.

### 3.1.2 Underwater image enhancement

For the case of underwater image enhancement, we utilized a modified function known as gain offset correction contrast stretching (GOC-CS) [10] to perform colour correction for the effect of water prior to enhancement. This configuration was compared with PDE-based formulations utilizing GOC1, GOC2, piecewise linear transform (PWL) and histogram specification (HS) [15].

### 3.1.3 Application to dust/sand-storm image enhancement

For dusty image enhancement, we utilize the same setup for underwater images since the images are degraded in a similar manner. The flowchart of the PDE-based formulation with the schemes for hazy/dust storm and underwater images is shown in Fig. 3. The colour enhancement version of the hazy image enhancement can be performed for faded images using a previously proposed algorithm called red-green-blue-intensity-value (RGB-IV) [23].

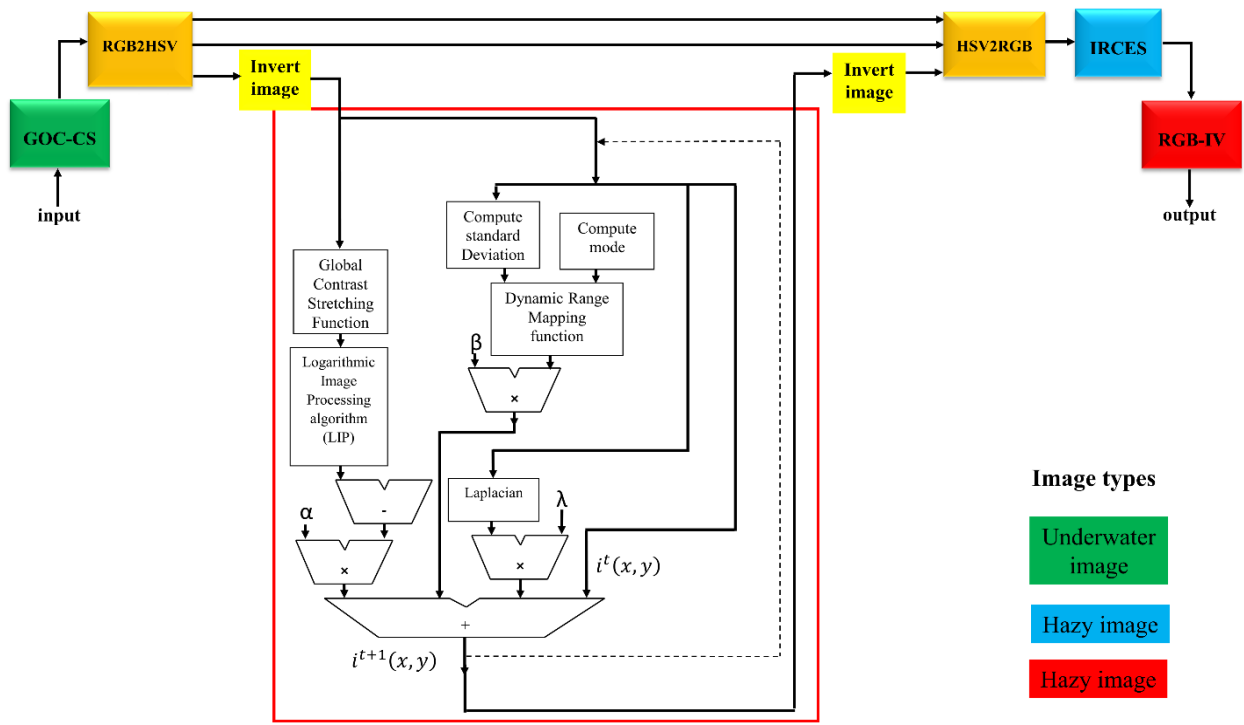

**Figure 3: Flowchart of PDE-based algorithm for hazy/underwater/dust storm image enhancement (PA)**

## 3.2 Fast de-hazing algorithm and modifications

In this section, we also propose a fast de-hazing algorithm using a simple approximation of the hazy image formation model. In the model a hazy image [24] is defined as shown;

$$I(x,y) = J(x,y).t(x,y) + A.[1 - t(x,y)] \quad (4)$$

In (4) $I(x,y)$ is the hazy image, $J(x,y)$ is the haze-free image or radiance, $A$ is the sky light [25] and $t(x,y)$ is the transmission map [25] expressed as;

$$t(x,y) = e^{-\int_0^{d(x,y)} \beta(r_{x,y}(s))ds} = e^{-\beta d(x,y)} \quad (5)$$

In (5), $\beta(.)$ is the scattering coefficient, $r_{x,y}$ is the light ray at pixel location, $x, y$ [25]. The objective of de-hazing process is to obtain $J(x,y)$ from $I(x,y)$ [1] as shown in (6);

$$J(x,y) = \frac{I(x,y) - A}{t(x,y)} + A \quad (6)$$

However, by noting from previous experiments that $t(x,y)$ is similar to the inverted image $I(x,y)$ or illumination component, $L(x,y)$, gives $t(x,y) \equiv 255 - L(x,y)$ or $t(x,y) \equiv 1 - L(x,y)$, if normalized. Additionally, **A** is equivalent to an array of ones or reflectance value and can be converted to a scalar value such that $A = 1$. The radiance, $J(x,y)$ becomes the de-hazed or enhanced output image. Thus we can rewrite this expression in (6) as;

$$J(x,y) = \frac{I(x,y)-\mathbf{1}}{1-L(x,y)} + \mathbf{1} \qquad (7)$$

Comparing (4) with the illumination-reflectance model; $I(x,y) = L(x,y).R(x,y)$ and assuming no airlight, **A**, eqn (4) reduces to;

$$I(x,y) = t(x,y).J(x,y) \equiv [1 - L(x,y)].J(x,y) \qquad (8)$$

Thus, the radiance, $J(x,y)$ is similar to the reflectance, $R(x,y)$, while $t(x,y)$ is the inverse of the illumination component as before and finding $J(x,y)$ is simpler in this case. We derive the equation for the alternate enhancement-based de-hazing algorithm as;

$$t'(x,y) = f\{t(x,y)\} \qquad (9)$$

In eqn. (9), $f\{t(x,y)\}$ is a local-global or multi-scale contrast enhancement function and $J(x,y)$ can be computed as;

$$J(x,y) = 255[1 - t'(x,y)] \qquad (10)$$

The global function used for $f\{t(x,y)\}$ is the LIP algorithm by Patrascu, while the local contrast function is the contrast limited adaptive histogram equalization (CLAHE).

The fast de-hazing algorithm initially resulted in clearly visible halos due to the CLAHE algorithm, which necessitated a global contrast function prior to the CLAHE (clip limit is set to 0.002 with tile size of 32). Reducing or increasing the tile size beyond this baseline leads to greater colour distortion and fading, which is worse around tile size of 8. By performing the LIP algorithm first, then followed by CLAHE, there is a smoother transition from low to high frequency regions as observed in the refined transmission maps of the de-hazed images. Increasing tile size above 32 to 64 leads to increase in high-pass filtering action, yielding more detail enhancement but distorted colours, halos and noise artifacts. However, in images with non-uniform or thick haze the algorithm leads to distorted colours. This occurs when fixing the clip limit at a particular value and adaptive computation does not lead to consistency.

Additionally, operating in the RGB colour space appears better than HSI or HSV space as the colours are seemed more consistent. Other colour space transformations lead to considerable colour fading. Based on experiments, the CLAHE introduces uncertainties and additional parameters to adjust. This makes this scheme unreliable and only suited to images with uniform and thin haze. Additional experiments were performed to fully render the algorithm adaptively; however, results were generally unsatisfactory. Thus, we reformulated the algorithm in the following form;

$$\bar{I}(x,y) = 255 - I(x,y) \qquad (11)$$

$$\bar{I}_{log}(x,y) = 20 log_{10}|\bar{I}(x,y)| \qquad (12)$$

$$t'(x,y) = f\{\bar{I}_{log}(x,y)\} = t_{BF}(x,y) \text{ or } t_{LPF}(x,y) \qquad (13)$$

In (11), $\bar{I}(x,y)$ is the inverted image. In eqn. (12), $\bar{I}_{log}(x,y)$ is the logarithm of the inverted image, $\bar{I}(x,y)$. In (13), the term, $f\{.\}$ can be a fractional order-based Gaussian low-pass ($t_{LPF}(x,y)$) or bilateral filtering ($t_{BF}(x,y)$) result. The former is chosen for speed. Then, the radiance or de-hazed image is obtained as;

$$J(x,y) = 255 \left[ \frac{I(x,y) - \bar{I}_{log}(x,y)}{255 - t'(x,y)} \right] \tag{14}$$

This is then incorporated into the existing PDE-based formulation to gradually improve the contrast. The results are shown in Fig. 4 and are better than the initial formulation with results more balanced than the previous modified PDE. This version does not require the CLAHE, thus maintains results free from halos and colour distortion and does not require tuning of multiple parameters. However, it also suffers from minimal local contrast enhancement and its run-time is similar to the modified PDE, which is slightly slower than the initial formulation.

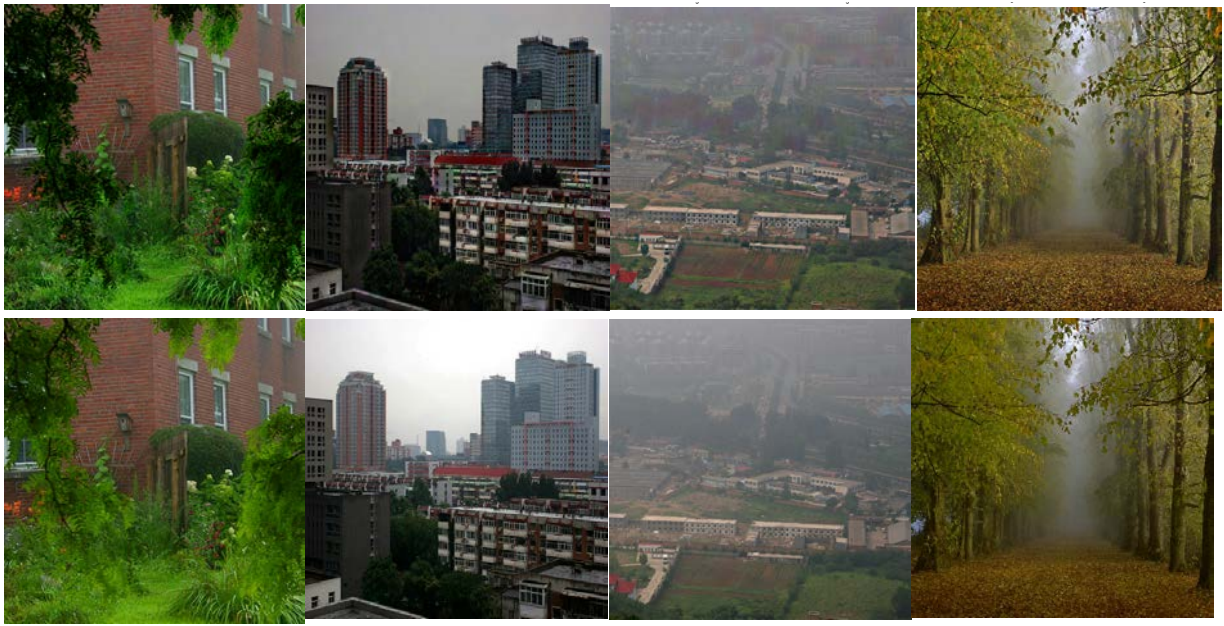

**Figure 4: Image results of initial PDE formulation (top row) and modified PDE formulation (bottom row)**

## 4  Experiments and results

Numerous experiments were performed to evaluate and compare PA with other methods using hazy, underwater and dusty images. Image quality measures utilized in assessing enhancement include entropy (E), (relative) average gradient (AG/RAG) [20], contrast enhancement factor (F) [26], global contrast factor (GCF) [27], colour enhancement measurement (EMEC) [28] and colourfulness/colour enhancement factor (CEF) [29]. Final values greater than the initial measurements imply improvement. The computing platform specifications are: Intel® Core i7-6500U x64-based 2.59 gigahertz (GHz) processor with 12 gigabyte (GB) random access memory (RAM) running 64-bit operating system (OS) and NVIDIA® GeForce™ 940M graphics processing unit (GPU) with compute capability of 5.0.

### 4.1  Hazy images

In Fig. 5, we present sample visual results from [6] amended with the results of PA for comparison. Based on visual assessment, PA yields comparable and brighter de-hazed results compared to several of the other algorithms. We verify this by utilizing the Fog Aware Density Evaluator (FADE) devised by Choi et al [30] and compare with other algorithms tested in work by Li and Zheng [6] as shown in Table 3. Lower FADE values indicate improved visibility and considerably reduced fog or haze density in the de-hazed images. Results show that PA yields comparable or best results for certain images despite being relatively less complex compared to the other approaches which involve the DCP or its variant.

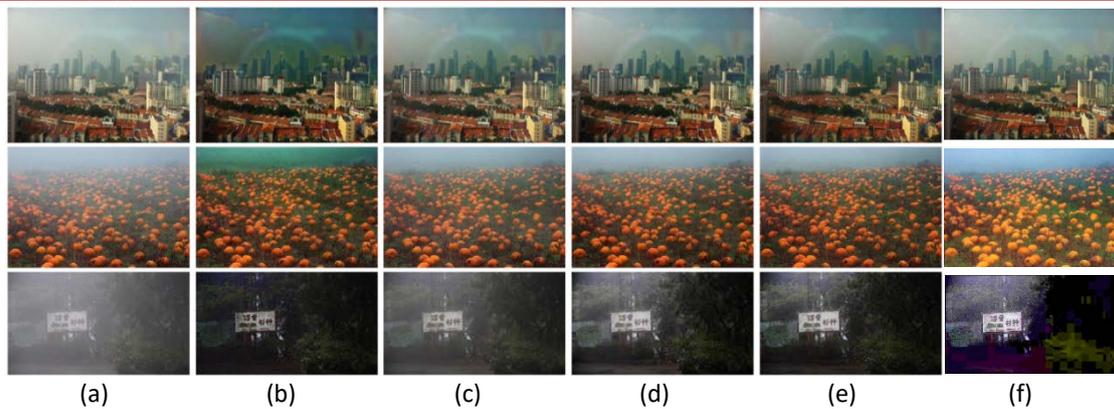

(a)      (b)      (c)      (d)      (e)      (f)

**Figure 5: (a) Original hazy images (b) He et al (c) Zhu et al (d) Li & Zheng1 (e) Li & Zheng2 (f) PA**

Table 2: Perceptual fog density comparison of PA with various de-hazing algorithms from [6]

| Algos / Image | He et al [3] | Zhu et al [31] | Li & Zheng1 [32] | Li & Zheng2 [6] | PA | Hazy image |
|---|---|---|---|---|---|---|
| *City2* | **0.255** | 0.424 | 0.354 | 0.301 | 0.975 | 0.658 |
| *Pumpkins* | **0.194** | 0.411 | 0.313 | 0.278 | 0.274 | 0.567 |
| *signpost* | 0.417 | 1.185 | 0.518 | 0.423 | **0.290** | 2.305 |
| *City3* | 0.992 | 3.293 | 1.402 | 0.937 | **0.593** | 7.447 |
| *City4* | 1.066 | 4.568 | 1.181 | 0.836 | 1.025 | 9.259 |
| *City5* | **0.399** | 0.835 | 0.535 | 0.437 | 0.583 | 2.28 |
| *Average* | 0.554 | 1.786 | 0.717 | **0.535** | 0.623 | 2.815 |

We also compare the methods by Ancuti and the DEnsity of Fog Assessment-based Defogger (DEFADE) algorithm proposed by Choi et al [30] to results of PA in Fig. 6. Visual results show that PA yields the sharpest most detailed image; however gradient reversal artifacts and darkened regions are observed for the building image, while the other images are more balanced and brighter than the results of Ancuti et al and the DEFADE algorithm. We also present FADE results from [30] amended with those obtained from images processed with PA for quantitative comparison in Table 3. The numerical values confirm the improved results of PA, which has the lowest haze density values with DEFADE yielding the second best results (red bolded).

We also compare with other algorithms such as those by Tan [33], Fattal [34], Kopf et al [35], He et al [3], Tarel et al [36] in addition to the method by Ancuti et al [37], DEFADE [30] and PA. The visual comparison in Fig. 6 is from Choi et al [30] amended with the image results of PA. The results of PA are shown without the brightening filter and are darker than the other results, though with increased contrast. The fog density values of the images processed with PA are compared with those of the various algorithms presented in [30] and shown in Table 5. Results indicate that PA yields images with the lowest perceptual fog density in most cases apart from the method by Tan [33].

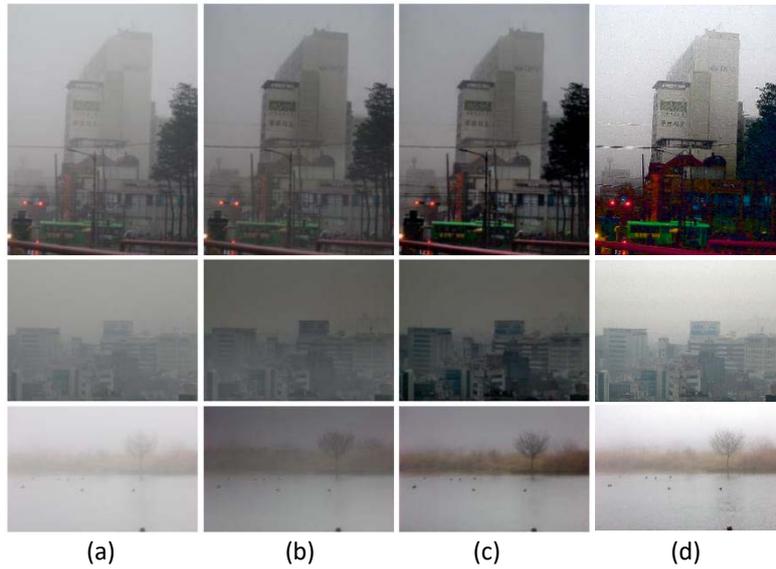

Figure 6: (a) Original hazy images (b) He et al (c) Zhu et al (d) Li & Zheng1 (e) Li & Zheng2 (f) PA

Table 3: Perceptual fog density comparison of PA with various de-hazing algorithms from [30]

| Algos Images | Ancuti et al | DEFADE | PA | Hazy image |
|---|---|---|---|---|
| *building* | 1.64 | **0.69** | **0.31** | 2.71 |
| *cityscape* | 1.73 | **0.85** | **0.70** | 3.46 |
| *river* | 2.64 | **1.75** | **1.06** | 5.77 |

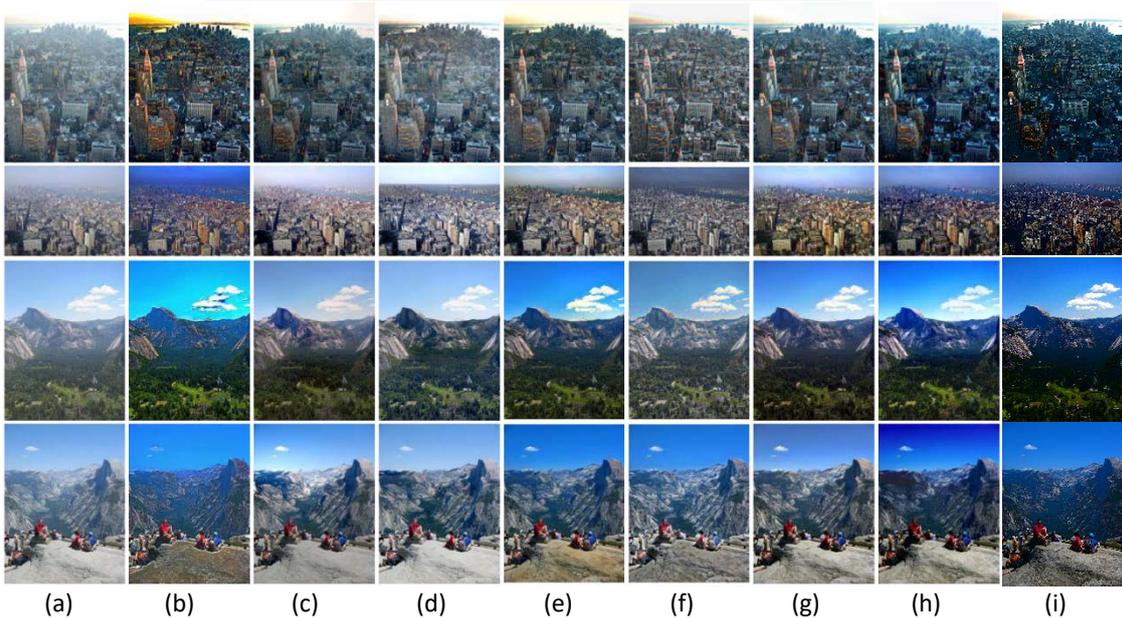

Figure 7: (a) Original hazy images (b) Tan [33] (c) Fattal [34] (d) Kopf et al [35] (e) He et al [3] (f) Tarel et al [36] (g) Ancuti et al [37] (h) DEFADE [30] (i) PA

Table 4: Perceptual fog density comparison of PA with various de-hazing algorithms from [30]

| Algos Image | Tan [33] | Fattal [34] | Kopf et al [35] | He et al [3] | Tarel et al [36] | Ancuti et al [37] | DEFADE [30] | PA | Hazy image |
|---|---|---|---|---|---|---|---|---|---|
| *Ny12* | **0.11** | 0.26 | 0.48 | 0.42 | 0.24 | 0.31 | 0.20 | **0.15** | 0.73 |
| *Ny17* | **0.17** | 0.36 | 0.41 | 0.27 | 0.24 | 0.30 | 0.20 | **0.14** | 0.67 |
| *Y01* | **0.18** | 0.51 | 0.51 | 0.29 | 0.31 | 0.33 | 0.26 | **0.23** | 0.91 |
| *Y16* | **0.21** | 0.43 | 0.48 | 0.27 | 0.25 | 0.33 | **0.24** | **0.21** | 0.63 |

## 4.2 Underwater images

We also compare the results of PA for underwater images with results of Khan et al [38] and Galdran et al [39]. Fig. 8 depicts the visual results from Khan et al [38] amended with PA for visual comparison. Visual observation indicates that the results of PA are the most vivid and colourful in most cases. However, some images are quite dark and in one particular image, distorted (*Ancuti2* image in penultimate image column, bottom row).

We also compare underwater image enhancement results of PA with the algorithms by He et al [3], Ancuti & Ancuti [37], Drews-Jr et al [40], Galdran et al [39], Emberton et al [41], Ancuti et al1 [7] and Ancuti et al2 [11]. The visual results are shown in Fig. 9, which is from [11] amended with results of PA. Once more, PA shows the most vivid results, though some images are dark and some colour distortion observed in the last image result of penultimate image row in Fig.9.

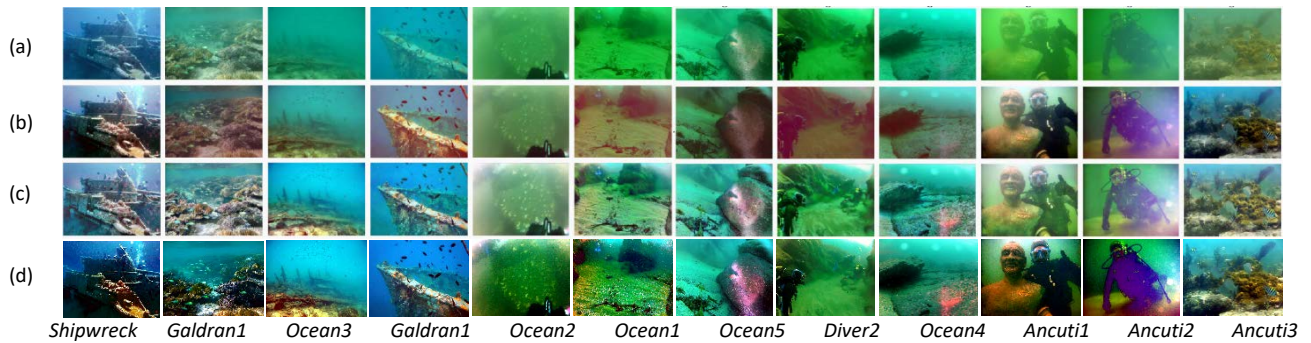

*Shipwreck   Galdran1   Ocean3   Galdran1   Ocean2   Ocean1   Ocean5   Diver2   Ocean4   Ancuti1   Ancuti2   Ancuti3*

**Figure 8: Image results from [38] amended with results of PA (a) Original underwater images (b) Galdran et al [39] (c) Khan et al [38] (d) PA**

We also quantitatively compare the algorithms using the Underwater Color Image Quality Evaluation (UCIQE) metric developed by Yang and Sowmya [42] in Table 5. Generally, the higher the numerical value of the UCIQE, the better the enhancement outcome. Numerical results show that PA consistently has the highest UCIQE values for almost all the processed images and the highest average UCIQE value. This is consistent with the considerable contrast and colour enhancement observed in most of the images processed with PA.

**Table 5: UCIQE value comparison of PA with various de-hazing and underwater enhancement algorithms from [38]**

| Algos \ Images | He et al [3] | Ancuti & Ancuti [37] | Drews-Jr [40] | Galdran et al [39] | Emberton et al [41] | Ancuti et al1 [7] | Ancuti et al 2 [11] | Khan et al [38] | PA |
|---|---|---|---|---|---|---|---|---|---|
| *Shipwreck* | 0.565 | 0.629 | 0.550 | **0.646** | 0.632 | 0.634 | 0.632 | 0.599 | **1.014** |
| *fish* | 0.602 | 0.650 | 0.623 | 0.527 | **0.705** | **0.669** | 0.667 | 0.625 | 0.616 |
| *Reef1* | 0.612 | 0.657 | 0.649 | 0.576 | 0.660 | 0.655 | 0.658 | **0.661** | **0.682** |
| *Reef2* | 0.702 | 0.683 | 0.659 | 0.633 | **0.718** | **0.718** | 0.711 | 0.681 | **0.722** |
| *Reef3* | 0.606 | 0.661 | 0.620 | 0.533 | 0.678 | **0.705** | 0.697 | 0.665 | **1.027** |
| *Galdran1* | 0.593 | 0.631 | 0.544 | 0.529 | 0.652 | 0.643 | **0.659** | 0.632 | **0.892** |
| *Galdran2* | 0.426 | 0.558 | 0.536 | 0.596 | 0.630 | **0.667** | 0.633 | 0.578 | **0.683** |
| *Ancuti1* | 0.485 | 0.561 | 0.499 | 0.641 | 0.499 | 0.558 | **0.594** | 0.542 | **0.935** |
| *Ancuti2* | 0.456 | **0.595** | 0.492 | 0.529 | 0.529 | 0.590 | 0.592 | 0.535 | **1.179** |
| *Ancuti3* | 0.577 | 0.643 | 0.535 | 0.614 | 0.555 | 0.652 | **0.664** | 0.609 | **0.784** |
| *Average* | 0.562 | 0.627 | 0.571 | 0.582 | 0.626 | 0.649 | **0.651** | 0.613 | **0.853** |

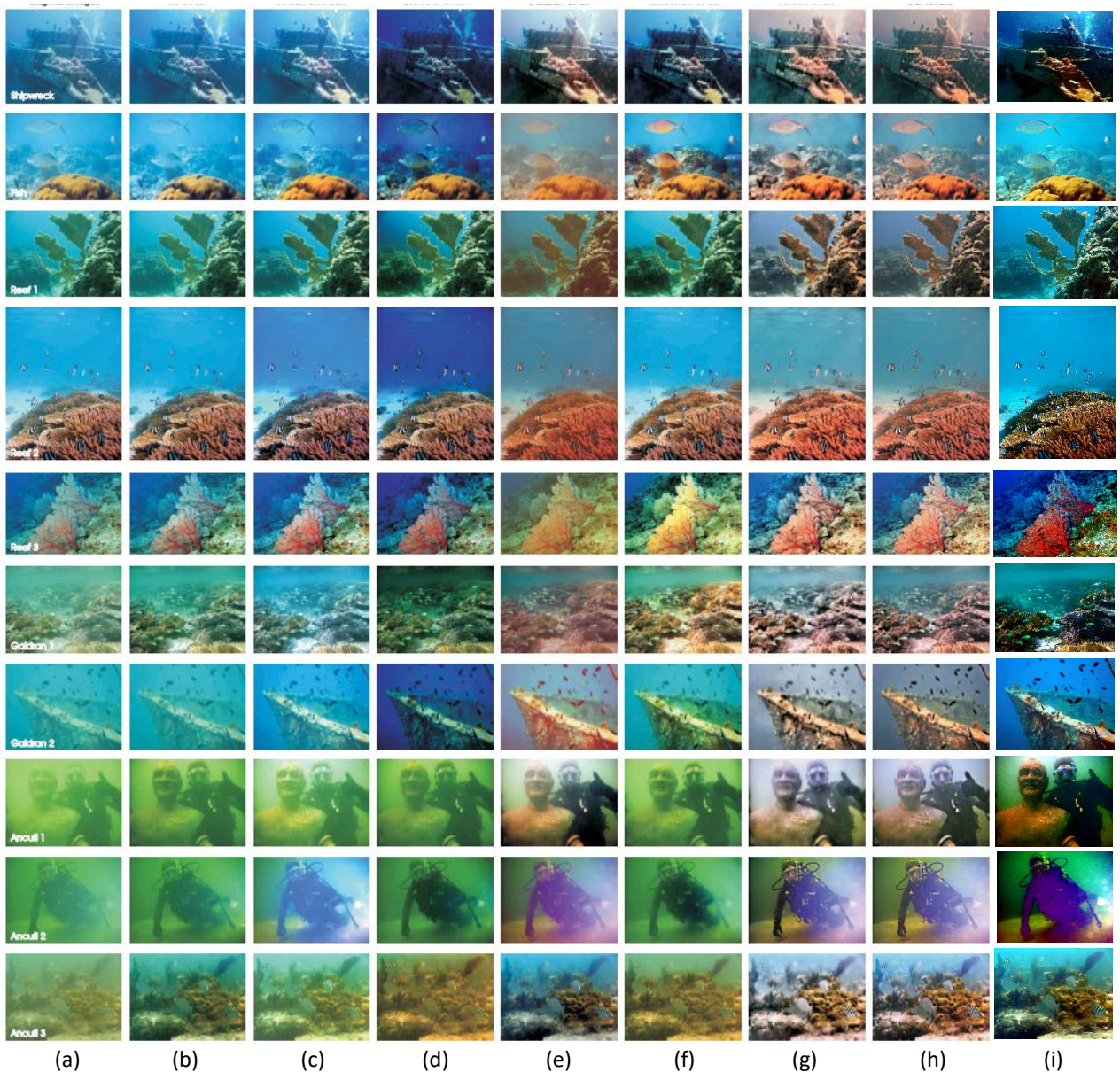

(a)   (b)   (c)   (d)   (e)   (f)   (g)   (h)   (i)

**Figure 9: Image results from [11] amended with results of PA (a) Original underwater images (b) He et al [3] (c) Ancuti & Ancuti [37] (d) Drews-Jr et al [40] (e) Galdran et al [39] (f) Emberton et al [41] (g) Ancuti et al1 [7] (h) Ancuti et al2 [11] (i) PA**

## 4.3 Sand/dust-storm images

For the sandstorm images, we use images from [43] and compare results with the proposed approach in Fig. 10 and Table 6. Results indicate that there is more contrast and colour enhancement using the proposed algorithm compared with the method by [43]. However, in some cases, the proposed approach leads to discolouration of processed images. These images have little overlap between red, green blue channel histograms, making colour correction much more difficult. However, such problems can be mitigated using previous approaches for such images, though this implies additional modifications. Based on results for hazy, underwater and dust storm images, PA is relatively versatile in handling these image categories with minimal modification, while problems of previous methods have been improved upon. However, the proposed approach still darkens certain images with some colour distortion. This is a shortcoming of the PA, which we address in the next subsection by substitution of the filter-based post enhancement operator with an edge-agnostic version.

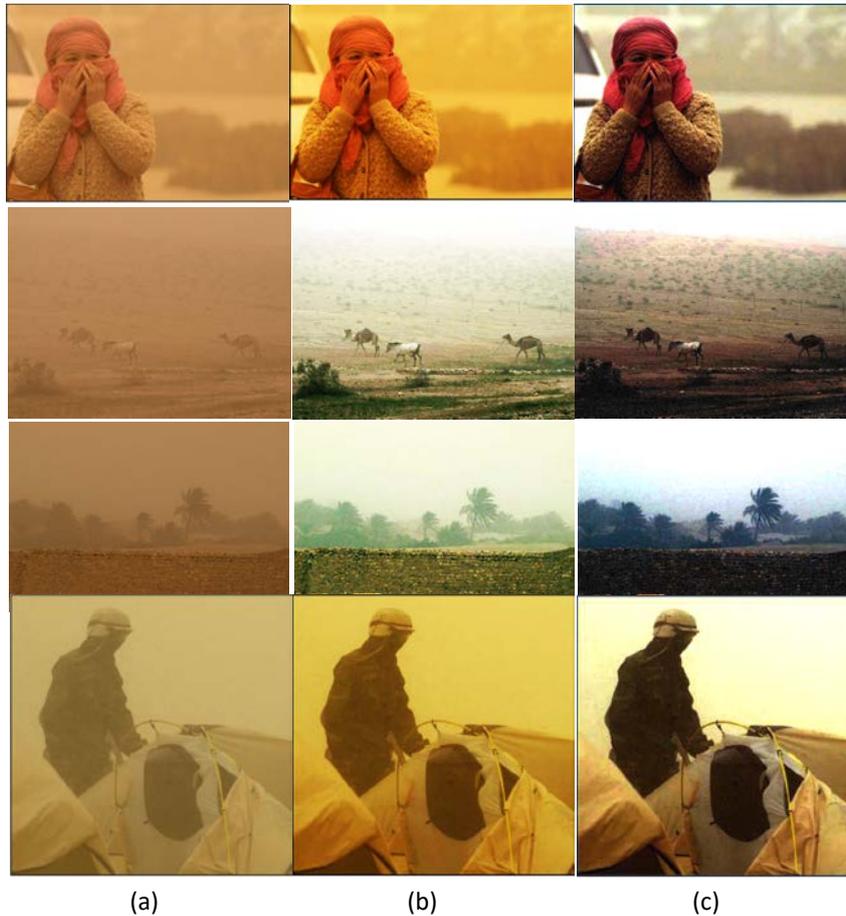

(a)            (b)            (c)

**Figure 10: (a) Original dust storm images [43] (b) Al-Ameen [43] (c) PA**

**Table 6: Quantitative comparison of PA with algorithm by Al-Ameen (AA) [43]**

| Images | CEF (PA/AA) | RAG (PA/AA) | E (PA/AA) | GCF (PA/AA) | F (PA/AA) | EMEC (PA/AA) |
|---|---|---|---|---|---|---|
| *Woman* | **1.2033**/ 0.9073 | **3.3956**/ 1.6968 | 7.2632/ **7.5746** | **6.8627**/ 4.0401 | **2.0459**/ 0.9867 | **48.5197**/ 11.0049 |
| *Camels* | **1.1683**/ 1.0419 | **14.5191**/ 4.4053 | **6.8365**/ 6.7967 | **6.4657**/ 3.7083 | **4.1640**/ 1.8591 | **48.2176**/ 5.6426 |
| *Trees* | 0.9946/ **1.1734** | **6.1000**/ 3.1496 | 5.0352/ **6.7707** | **7.1997**/ 4.1148 | **4.0329**/ 1.7449 | **46.2202**/ 9.3487 |
| *Rider* | 1.7663/ **1.8642** | **3.1853**/ 1.3990 | 6.1114/ **7.4213** | **6.3387**/ 3.7311 | **3.3965**/ 2.4691 | **57.4407**/ 16.5225 |

## 4.4 Rectification of darkening of image results using fuzzy homomorphic enhancements

We wish to resolve the darkening of the processed images for hazy, dust and underwater images and require a low-complexity, edge-agnostic and effective tonal correction algorithm. Thus, we utilize the fuzzy homomorphic enhancement (FHE) algorithm from previous work [16] to replace the IRCES as a post processing step in PA.

The improvements are shown (and compared with previous results) in Fig. 11 for hazy, underwater and dust storm images. The modification of the algorithm does not affect the other processed images, which are not darkened by the de-hazing process. Thus, we have rectified the issue of darkened images, while maintaining good de-hazing outcomes. Also, this modification reduces the runtime of PA when the FHE is utilized in the PDE formulation, eliminating the need for a post-processing. This also results in faster convergence of the PDE and a more balanced enhanced output free from edge artifacts.

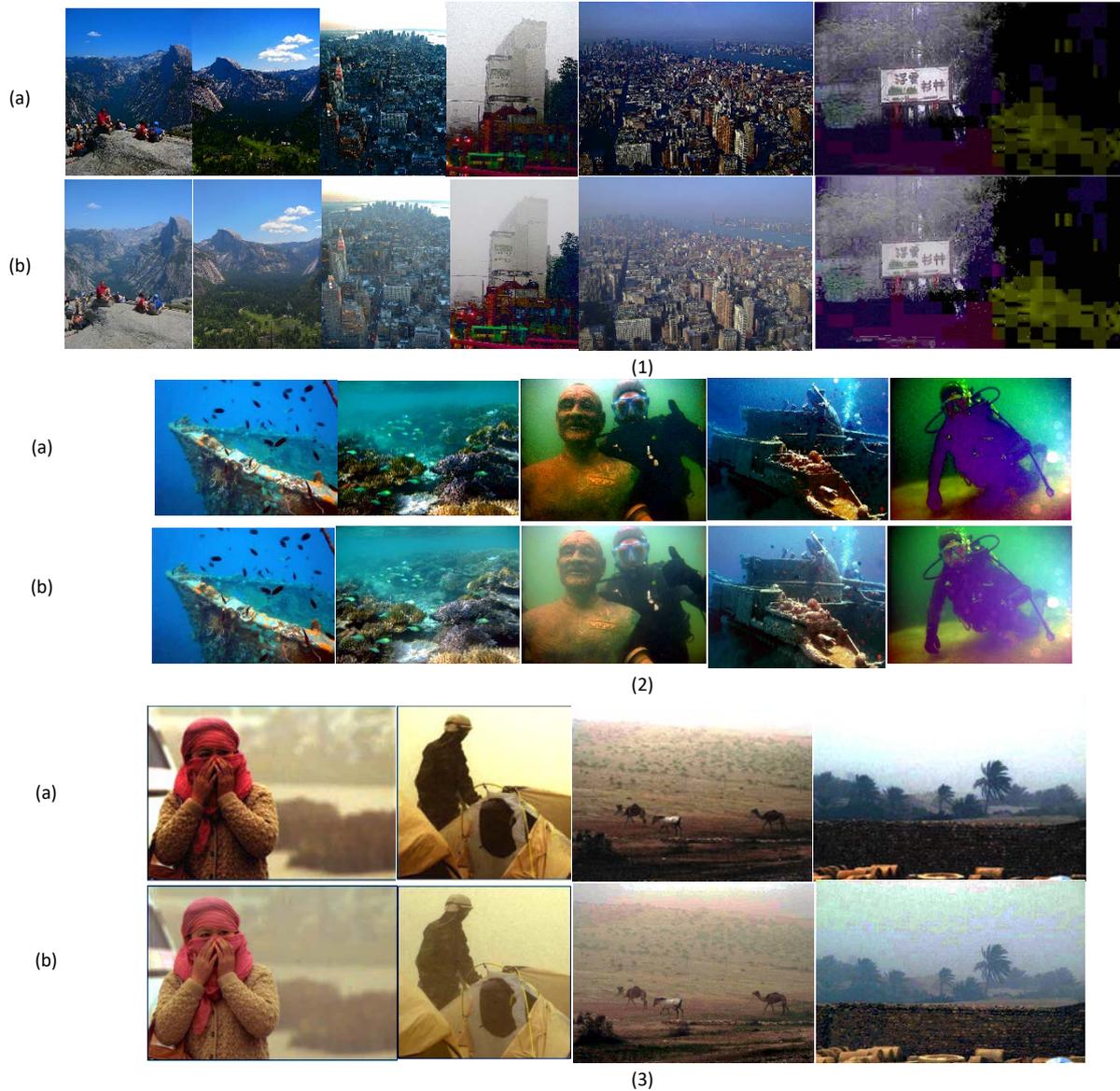

**Figure 11: (a) Initial enhanced and (b) brightened (1) hazy (2) underwater (3) dust storm images using PA**

## 5  Conclusion

This paper has presented the theoretical formulation and experimental validation of modified PDE-based image enhancement algorithms for hazy and underwater images. Problems such as darkening, sky region, excessive edge and noise enhancement were addressed by utilizing a soft edge-agnostic LIP algorithm and utilizing the spatial filter as post-processing enhancement operation. Thus, absence of halo effect and sky region enhancement is maintained in the result of the modified PDE formulations in addition to brightening of dark regions. A fuzzy homomorphic enhancement algorithm, which rectified the darkening of certain outlier images was utilized as a post-processing alternative to the IRCES. Moreover, the proposed modified algorithms have been employed in underwater and dust storm image enhancement with mostly improved results compared to several well-known and more complex algorithms from the literature.

**REFERENCES**

[1]. Sungmin Lee, Seokmin Yun, Ju-Hun Nam, Chee Sun Won, and Seung-Won Jung, *A review on dark channel prior based image dehazing algorithms*. EURASIP Journal on Image and Video Processing, vol. 2016, no. 4, pp. 1-23, 2016..

[2]. Dilbad Singh and Vijay Kumar, *A Comprehensive Review of Computational Dehazing Techniques*. Archives of Computational Methods in Engineering, pp. 1-13, September 2018.

[3]. Kaimin He, Jian Sun, and Xiaoou Tang, *Single Image Haze Removal Using Dark Channel Prior*. IEEE Transactions on Pattern Analysis and Machine Intelligence (PAMI), vol. 33, no. 12, pp. 2341-2353, 2010.

[4]. Bo Li, Shuhang Wang, Jin Zheng, and Liping Zheng, *Single image haze removal using content-adaptive dark channel and post enhancement*. IET Computer Vision, vol. 8, no. 2, pp. 131-140, April 3 2014.

[5]. Kaiming He, Jian Sun, and Xiaoou Tang, *Guided image filtering*. IEEE transactions on pattern analysis and machine intelligence, vol. 35, no. 6, pp. 1397-1409, June 2013.

[6]. Zhengguo Li and Jinghong Zheng, *Single Image De-Hazing Using Globally Guided Image Filtering*. IEEE Transactions on Image Processing, vol. 27, no. 1, pp. 442-450, 8 September 2017.

[7]. C. Ancuti, CO. Ancuti, T. Haber, and P. Bekaert, *Enhancing underwater images and videos by fusion*. IEEE Conference on Computer Vision and Pattern Recognition, Jun 16 2012, pp. 81-88.

[8]. Adrian Galdran, Javier Vazquez-Corral, David Pardo, and Marcelo Bertalmıo, *Fusion-based Variational Image Dehazing*. IEEE Signal Processing Letters, vol. 24, no. 2, pp. 151-155, Feb 2017.

[9]. Adrian Galdran, *Artificial Multiple Exposure Image Dehazing*. Signal Processing, vol. 149, pp. 135-147, August 2018.

[10]. U. A. Nnolim, *Adaptive Multi-Scale Entropy Fusion De-Hazing Based on Fractional Order*. Journal of Imaging, vol. 4, no. 9, p. 108, September 6 2018.

[11]. Codruta O. Ancuti, Cosmin Ancuti, Christophe De Vleeschouwer, and Philippe Bekaert, *Color Balance and Fusion for Underwater Image Enhancement*. IEEE Transactions on Image Processing, vol. 27, no. 1, pp. 379-393, January 2018.

[12]. Wenqi Ren et al., *Single Image Dehazing via Multi-Scale Convolutional Neural Networks*. European Conference on Computer Vision, Springer International Publishing, Oct. 8, 2016 , pp. 154-169.

[13]. R. C. Gonzalez and R. E. Woods, *Digital Image Processing, 2nd edition*.: Prentice Hall, 2002.

[14]. Vicent Caselles, Jean-Michel Morel, Guillermo Sapiro, and Allen Tannenbaum, *Introduction to the Special Issue on Partial Differential Equations and Geometry-Driven Diffusion in Image Processing and Analysis*. IEEE Transactions on Image Processing, vol. 7, no. 3, pp. 269-273, March 1998.